\documentclass[conference]{IEEEtran}
\IEEEoverridecommandlockouts
\usepackage{float}
\usepackage{placeins}
\usepackage{cite}
\usepackage{amsmath,amssymb,amsfonts}
\usepackage{algorithmic}
\usepackage{graphicx}
\usepackage{textcomp}
\usepackage{subcaption}  
\usepackage{booktabs}
\usepackage{caption}     
\usepackage{xcolor}
\usepackage{url}
\usepackage{hyperref}    
\hypersetup{
    colorlinks=true,
    linkcolor=green,
    citecolor=green,
    urlcolor=green
}
\def\BibTeX{{\rm B\kern-.05em{\sc i\kern-.025em b}\kern-.08em
    T\kern-.1667em\lower.7ex\hbox{E}\kern-.125emX}}
\begin{document}

\title{Multi-Class Parkinson’s Disease Detection Based on Finger Tapping Using Attention-Enhanced CNN–BiLSTM}
\author{
    \IEEEauthorblockN{1\textsuperscript{st} Abu Saleh Musa Miah}
     \IEEEauthorblockN{3\textsuperscript{rd} Najmul Hassan}
    \IEEEauthorblockA{
          \textit{School of Computer Science and Engineering} \\
        \textit{University of Aizu}, Fukushima, Japan \\
      abusalehcse.ru@gmail.com
    }
    \and
    \IEEEauthorblockN{2\textsuperscript{nd} Md Maruf Al Hossain}
     \IEEEauthorblockN{4\textsuperscript{th} Ayesha Siddiqua}
    \IEEEauthorblockA{
        \textit{Department of Computer Science and Engineering} \\
        \textit{Army University of Science and Technology}, Nilphamari, Bangladesh \\
       }
   
    \and
    \IEEEauthorblockN{5\textsuperscript{th} Yuichi Okuyama }
    \IEEEauthorblockA{
        \textit{School of Computer Science and Engineering} \\
        \textit{University of Aizu}, Fukushima, Japan \\
       Okuyama@u-aizu.ac.jp
         }
    \and
    \IEEEauthorblockN{6\textsuperscript{th} Jungpil Shin*}
    \IEEEauthorblockA{
        \textit{School of Computer Science and Engineering} \\
        \textit{University of Aizu}, Fukushima, Japan \\
        jpshin@u-aizu.ac.jp}}
\maketitle
\begin{abstract}
Accurate evaluation of Parkinson's disease (PD) severity is essential for effective clinical management and intervention development. Despite the proposal of several gesture-based PD recognition systems, including those using the finger-tapping task to assess Parkinsonian symptoms, their performance remains unsatisfactory. In this study, we present a multi-class PD detection system based on finger-tapping, using an attention-enhanced CNN–BiLSTM framework combined with handcrafted feature extraction and deep learning techniques. In the procedure, we used an existing dataset of finger-tapping videos to extract temporal, frequency, and amplitude-based features from wrist and hand movements using their formulas. These handcrafted features were then processed through our attention-enhanced CNN–BiLSTM model, a hybrid deep learning framework that integrates CNN, BiLSTM, and attention mechanisms to classify PD severity into multiple levels. The features first pass through a Conv1D–MaxPooling block to capture local spatial dependencies, followed by processing through a BiLSTM layer to model the temporal dynamics of the motion. An attention mechanism is applied to emphasize the most informative temporal features, which are then refined by a second BiLSTM layer.
The CNN-derived features and attention-enhanced BiLSTM outputs are concatenated, followed by dense and dropout layers, before being passed through a softmax classifier to predict the PD severity level. Our model demonstrated strong performance in distinguishing between the five severity classes, showcasing the effectiveness of combining spatial-temporal representations with attention mechanisms for automated PD severity detection. This approach offers a promising non-invasive tool to assist clinicians in monitoring PD progression and making informed treatment decisions.
\end{abstract}
\section{Introduction}
\label{sec:introduction}
Parkinson's disease (PD) is a progressive neurodegenerative disorder affecting over 10 million people worldwide, marked by motor symptoms such as tremor, rigidity, bradykinesia, and postural instability \cite{islam2023using,joseph2023parkinson,hasan2024hi5,miah2025methodologica_pd,shin2025autism_miah,matsumoto2025machine_miah_pd,shin2025parkinson}. PD severity is commonly assessed using rating scales like the Unified Parkinson's Disease Rating Scale (UPDRS) and its MDS-sponsored revision (MDS-UPDRS) \cite{martinez-martin2016analysis,shin2025parkinson}. While clinically valuable, these assessments are subjective, time-consuming, and prone to inter-rater variability, which can reduce consistency in tracking disease progression and treatment outcomes \cite{capecci2019kimore,islam2023using}.
The need for precise, objective, and accessible PD severity assessment has fueled interest in automated diagnostics. Digital biomarkers from wearable sensors, smartphone apps, and video analysis offer alternatives to conventional evaluations \cite{sobolev2021advancing,carissimo2022development}. Video-based methods are particularly attractive for their non-invasive nature, accessibility, and ability to capture full-body movement without specialized hardware \cite{dunn2022application}. Advances in computer vision and machine learning have enabled automated movement analysis, with traditional ML approaches, such as using handcrafted features, showing strong potential for PD detection and severity classification \cite{sotirakis2023identification, suquilanda-pesantez2020prediction, tong2021severity}. Various machine learning algorithms, including SVMs, decision trees, and random forests, have been applied to develop Parkinson's disease detection systems \cite{trabassi2022machine, zhang2022mining, Alessia_Sarica_2022}. However, these methods often rely on domain expertise for feature engineering and may not fully capture the complex temporal dynamics and subtle movement patterns characteristic of different PD severity levels. Recently, there have been many researchers working to develop a PD recognition system using handwriting \cite{shin2025parkinson}, and video-based gestures such as postural stability, toe tapping, and finger tapping, etc. Matsumoto et al. recently developed a finger-tapping video-based Parkinson's Disease (PD) recognition system using various machine learning algorithms \cite{matsumoto2025machine_miah_pd}. Building on this, Islam et al. introduced a new finger-tapping video dataset and proposed a set of handcrafted feature extraction techniques combined with diverse machine learning algorithms for PD recognition. Their analysis showed a mean absolute error (MAE) of 0.58 points, compared to the raters' average MAE of 0.83 points, though the model performed slightly worse than expert neurologists, who achieved an MAE of 0.53 points. Despite these promising results, their performance metrics, such as accuracy, remained below 60\%, highlighting the challenge of achieving high performance for finger-tapping-based PD recognition, especially for real-life deployment.
In recent years, deep learning approaches have shown outstanding performance across various domains \cite{al-qurashi2020recurrent, lemieux2020hierarchical}. Convolutional Neural Networks (CNNs) are effective at capturing local spatial patterns in biomedical data, while Recurrent Neural Networks (RNNs), particularly Long Short-Term Memory (LSTM) networks, are adept at modeling temporal dependencies in sequential inputs \cite{wang2022lstm, zuo2015convolutional, hernandez2021attention, tang2018analysis}. However, few researchers have applied deep learning to finger-tapping-based PD recognition.

In this study, we build upon previous finger-tapping video datasets and their handcrafted features \cite{islam2023using} by proposing an attention-enhanced CNN-BiLSTM model. This hybrid model combines CNN, BiLSTM, and attention mechanisms, allowing it to effectively leverage both spatial and temporal features. The integration of attention mechanisms enhances interpretability and performance by focusing on the most relevant input elements, further improving the model’s ability to classify PD severity accurately. This approach not only improves the recognition performance but also enhances the potential for real-world deployment in clinical settings, addressing the challenges faced by earlier methods.


\section{Related works}\label{sec1}
Many researchers are working on PD recognition using machine learning and deep learning algorithms \cite{jibon2024parkinson, islam2024review}. 

Deng et al. used video-based systems to capture kinematic data to identify key movement features—such as pinkie finger motion and gait parameters—linked to PD severity via MDS-UPDRS Part III \cite{deng2024interpretable}. These rich kinematic features enable algorithms to detect subtle patterns indicating disease progression. Another researcher, Sotirakis et al., used wearable sensors to complement video analysis by providing continuous, high-resolution data. Machine learning applied to these features can outperform traditional clinical scales in detecting motor changes, offering greater sensitivity for monitoring progression \cite{sotirakis2023identification}. In the same ways, classical algorithms like SVMs, decision trees, and random forests used to classify PD based on gait features from trunk acceleration \cite{trabassi2022machine}, aided by feature selection methods that improve interpretability and accuracy. Multi-modal integration of imaging and clinical data enhances diagnostic performance, reducing errors and enabling earlier intervention through complementary data sources \cite{zhang2022mining}. The Explainable Boosting Machine (EBM) combines clinical and imaging features to classify PD and SWEDD with high AUC-ROC, offering transparent global and local explanations that enhance clinician trust \cite{Alessia_Sarica_2022}.

As the video dataset is usually large size and it's difficult to handle a large dataset using ML algorithms then researchers focus on the deep learning algorithm. Cui et al. applied the FResnet18 model, which fuses handcrafted texture features (Local Binary Pattern, Gray-Level Co-occurrence Matrix) with deep MRI features from a modified ResNet18, achieving 98.66\% accuracy \cite{cui2022diagnosis}. Similarly, Anitha et al. applied AttentionLUNet, which mainly integrates LeNet and U-Net with attention and residual modules, reaching 99.58\% accuracy on MRI data and highlighting the importance of preprocessing and augmentation \cite{palakayala2024attentionlunet}. More recently, the integration of Convolutional Neural Networks (CNN) and Long Short-Term Memory (LSTM) offers a powerful framework for PD assessment, combining spatial feature extraction, temporal modeling, and selective focus. CNNs effectively capture spatial hierarchies, as shown by a CNN-LSTM model using Mel-spectrograms for speech-based PD detection, achieving 93.51\% accuracy \cite{lilhore2023retracted}. LSTMs excel at modeling temporal dependencies, with 3D multi-head attention residual networks successfully identifying subtle movement \cite{shin2025multimodal,shin2025fall} variations for PD severity evaluation \cite{huang2024parkinson}. Attention mechanisms, such as in the ASGCNN model, enhance diagnostic accuracy (87.67\%) by focusing on the most relevant EEG channels \cite{chang2023eeg}, offering clinically valuable insights into disease mechanisms and progression.
Despite advances in deep learning for PD detection, several challenges hinder clinical adoption. A key limitation is the lack of interpretability, which reduces clinician trust even when models achieve state-of-the-art accuracy \cite{skaramagkas2023multi}. This is critical given PD’s symptom overlap with other neurological disorders \cite{islam2025diabd,hassan2025neurological}, making transparent decision-making essential for differential diagnosis \cite{lin2022early}.
\begin{figure*}[tb]
\centering
\includegraphics[width=7.5  in]{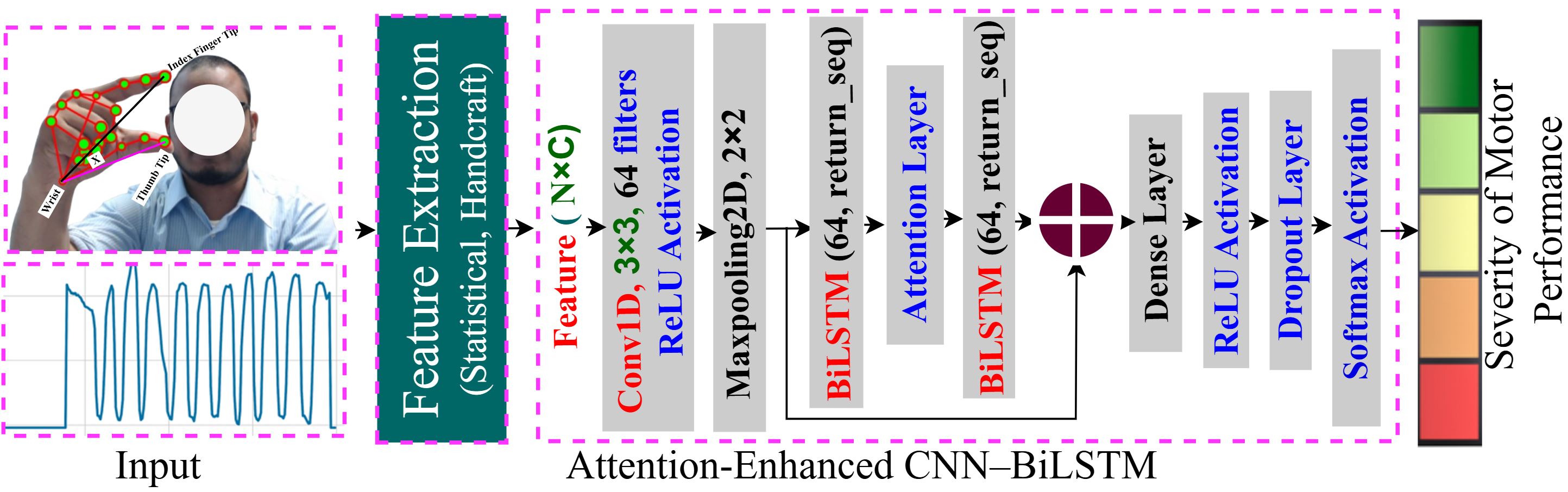}
\caption{Proposed deep learning based architecture for PD detection \cite{islam2023using}.}
\label{fig:main_diagram}
\end{figure*}

\begin{figure*}[tb]
    \centering
    \includegraphics[width=1\linewidth]{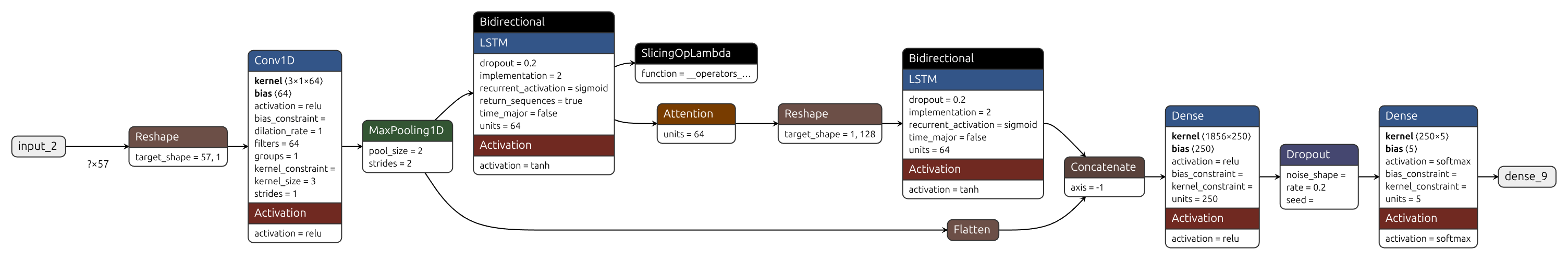}
    \caption{Details of the proposed Attention Enhanced CNN-BiLSTM architecture}
    \label{fig:model_arch}
\end{figure*}


\section{Proposed model} \label{sec3}
In this study, we propose a multi-class Parkinson’s Disease (PD) detection system based on finger-tapping, utilizing an attention-enhanced CNN–BiLSTM framework combined with handcrafted feature extraction and deep learning techniques. Using an existing dataset of finger-tapping videos, we extracted temporal, frequency, and amplitude-based features from wrist and hand movements, as described in \cite{islam2023using}. These handcrafted features were then processed through our hybrid deep learning model, which integrates Convolutional Neural Networks (CNN), Bidirectional Long Short-Term Memory (BiLSTM), and an attention mechanism to classify PD severity into five levels. The features are first reshaped to match the dimensions required for convolutional operations. A Conv1D layer is applied, followed by a MaxPooling1D layer to capture local spatial dependencies and reduce the dimensionality of the feature map. The output from the CNN layers is passed through a BiLSTM layer to model the temporal dynamics of the motion, capturing dependencies in both directions to account for both past and future context. To prevent overfitting, the BiLSTM layer is configured with a dropout rate of 0.2, enhancing the model’s generalization ability.
An attention mechanism is introduced to focus on the most informative temporal features. The attention layer computes attention weights by combining the outputs of the BiLSTM layer with the final hidden state, generating a context vector that highlights the most relevant sequence parts. This context vector is then processed through another BiLSTM layer to extract higher-level features. The final features from the CNN, BiLSTM, and attention modules are concatenated into a combined feature vector, which is processed through a Dense layer with 250 units and a Dropout layer (with a rate of 0.2) for regularization. The output is passed through a Dense layer with a softmax activation function to classify the input into one of five PD severity categories. The model is trained using sparse categorical cross-entropy as the loss function, and the Adam optimizer is used for efficient gradient-based optimization. The model’s performance is evaluated based on accuracy, and the architecture is summarized using model summary output for verification.
\subsection{Dataset and Clinical Ratings}
\textcolor{black}{
In this study, we used the publicly available ParkTest dataset, which includes finger-tapping task videos and corresponding clinical severity ratings based on the MDS-UPDRS criteria. The dataset was originally collected and rated by a team of expert neurologists and certified independent raters, as described in \cite{islam2023using}. The videos were collected from 250 global participants who recorded themselves completing the finger-tapping task in front of a computer webcam. Data were primarily collected at participants’ homes; however, a group of 48 individuals completed the task in a clinic using the same web-based tool. No additional annotation or re-rating was performed. This approach allows us to build upon a rigorously validated dataset and focus on the development and evaluation of our proposed model.}.
\subsection{Feature Extraction}
\textcolor{black}{
In this study, we reused the publicly available finger-tapping video dataset and corresponding clinical severity ratings originally provided by Islam et al. \cite{islam2023using}. The dataset includes videos collected via a web-based tool and annotated by both expert neurologists and MDS-UPDRS-certified raters following standardized protocols. For our analysis, we used their expert consensus ratings as the ground truth. We directly utilized the temporal features provided in the original study—including finger-tapping speed, acceleration, frequency, period, amplitude, and wrist displacement—without performing additional feature extraction. These features were derived from 2D hand key points using Google’s MediaPipe Hands framework by Islam et al. \cite{islam2023using}. The method includes steps such as hand detection, separation of left and right hand sequences, angle computation from specific hand landmarks (WRIST, THUMB\_TIP, and INDEX\_FINGER\_TIP), and post-processing techniques to handle noise, interpolate missing values, and isolate clean segments for analysis. Temporal features, including finger-tapping speed, acceleration, frequency, period, amplitude, and wrist displacement, were extracted and used in our model without modification. This approach enables consistent benchmarking and fair comparison with prior work while reducing preprocessing complexity in our proposed system.}

\subsection{Deep Learning Based Classification}
In this section, we describe the deep learning-based model for PD detection, including the mathematical derivation of the attention mechanism and its integration with the Convolutional Neural Network (CNN) and Bidirectional Long Short-Term Memory (BiLSTM) layers \cite{blake2025detection}.

\subsubsection{Input Representation}
Let the input feature matrix be denoted as:
\[
X = \{x_1, x_2, \ldots, x_N\}
\]
where \(N\) is the number of samples, and \(x_i \in \mathbb{R}^{57}\) represents the feature vector for the \(i\)-th sample. The feature vector consists of 57 individual features extracted from the raw data.

\subsubsection{Reshaping the Input}
Before passing the data through the convolutional layers, the input is reshaped into a sequence format for compatibility with the \texttt{Conv1D} layer:
\[
X_{\text{reshaped}} = \text{Reshape}(X) \in \mathbb{R}^{N \times 57 \times 1}
\]
where each sequence has a length of 57, and each feature is treated as a single channel in the Conv1D layer.

\subsubsection{Convolutional Layer}
The input sequence is then processed using a \texttt{1D convolutional layer} with 64 filters and a kernel size of 3. The output of this layer is:
\[
X_{\text{conv}} = \text{Conv1D}(X_{\text{reshaped}})
\]
This operation is followed by a \texttt{MaxPooling1D} layer that reduces the dimensionality of the sequence:
\[
X_{\text{pool}} = \text{MaxPooling1D}(X_{\text{conv}})
\]
The pooling operation retains the most significant features while reducing the feature map size.

\subsubsection{Bidirectional LSTM Layer}
The output from the convolutional and pooling layers is then passed through a Bidirectional LSTM layer, which captures temporal dependencies in both directions \cite{blake2025detection}. The BiLSTM layer is applied as:
\[
h_t = \text{BiLSTM}(X_{\text{pool}})
\]
where \(h_t \in \mathbb{R}^{64}\) represents the hidden state at time step \(t\) in the BiLSTM layer. The BiLSTM combines the outputs of both forward and backward passes, allowing the model to consider both past and future context.

\subsubsection{Attention Mechanism}
The attention mechanism is applied to the output of the BiLSTM to focus on the most important features for PD classification. The attention mechanism can be mathematically formulated as follows:
- The attention score is computed by applying two dense layers, \(W_1\) and \(W_2\), and a tanh activation function:
\[
\text{score}(i,j) = \tanh(W_1 h_i + W_2 h_j)
\]
where \(h_i\) and \(h_j\) are the hidden states at time step \(i\) and \(j\), and \(W_1, W_2\) are learnable weight matrices. The attention score is passed through another dense layer \(V\) to obtain the attention weights:
\[
\alpha_{ij} = \text{softmax}(V(\text{score}(i,j)))
\]
where \(\alpha_{ij}\) represents the attention weight for the \(i\)-th feature with respect to the \(j\)-th hidden state. The attention weights are normalized using the \texttt{softmax} function:
\[
\alpha_{ij} = \frac{e^{\text{score}(i,j)}}{\sum_{i=1}^n e^{\text{score}(i,j)}}
\]
- The context vector is computed as a weighted sum of the features, using the attention weights:
\[
c_j = \sum_{i=1}^{n} \alpha_{ij} h_i
\]
where \(c_j\) is the context vector that focuses on the relevant parts of the sequence. This context vector is reshaped and passed through an additional \texttt{Bidirectional LSTM} layer to extract higher-level features.

\subsubsection{Combining Features}

After obtaining the context vector from the attention layer, the CNN and LSTM features are concatenated to form a combined feature vector:

\[
X_{\text{combined}} = [\text{Flatten}(X_{\text{conv}}), \text{Flatten}(h_{\text{final}})]
\]

where \(\text{Flatten}(X_{\text{conv}})\) represents the flattened output of the convolutional layer and \(h_{\text{final}}\) is the final hidden state from the second BiLSTM layer.

\subsubsection{Fully Connected and Output Layers}

The combined feature vector is passed through a \texttt{Dense layer} with 250 units and a \texttt{Dropout} layer to prevent overfitting. The output layer uses the \texttt{softmax activation} function for classification into one of five categories of PD:

\[
\text{preds} = \text{Dense}(X_{\text{combined}})
\]

The output predictions are computed using:

\[
\hat{y} = \text{softmax}(\text{preds})
\]

where \(\hat{y} \in \mathbb{R}^5\) represents the predicted class probabilities.

\section{Experimental Result}
The proposed hybrid CNN–BiLSTM attention model achieved an overall accuracy of 93\% on the test set, demonstrating superior performance compared to the baseline approach by Islam et al. \cite{islam2023using}. The model was trained efficiently, requiring only 31.82 seconds for 100 epochs on the gait feature dataset.
Table \ref{tab:classification_metrics} presents the detailed classification metrics for each severity class. The model performed well across all classes, with macro average precision, recall, and F1-score of 95.40\%, 93.40\%, and 94.20\%, respectively. Class 4 (severe) achieved perfect precision, recall and F1-score 100.00\%, while Class 2 showed strong recall 97.00\% with a precision of 90.00\%. The macro-average F1-score of 94.20\% indicates robust performance across all severity levels.
Compared to the baseline model \cite{islam2023using}, our approach shows significant improvements in classification accuracy and correlation with ground truth severity scores. The confusion matrix analysis reveals excellent diagonal concentration with minimal misclassification, particularly for severe cases (Classes 3-4), which is clinically significant for Parkinson's disease assessment. The confusion matrix is shown in Fig. \ref{fig:conf_mat}. The attention mechanism successfully captured relevant temporal patterns in gait features, with the model showing consistent performance across different severity levels. The hybrid architecture effectively combined spatial feature extraction through CNN layers with temporal dependency modeling via BiLSTM, resulting in improved discrimination between adjacent severity classes.

\begin{table}[tb]
\centering
\caption{Classification Report (without support column)}
\begin{tabular}{l|c|c|c|}
\hline
Class & Precision & Recall & F1-Score \\
\hline
0 & 95.00 & 95.00 & 95.00 \\
1 & 92.00 & 92.00 & 92.00 \\
2 & 90.00 & 97.00 & 93.00 \\
3 & 100.00 & 83.00 & 91.00 \\
4 & 100.00 & 100.00 & 100.00 \\
\hline
Macro Avg & 95.40 & 93.40 & 94.20 \\
\hline
Accuracy &  &  & 93.00 \\
\hline
\end{tabular}
\label{tab:classification_metrics}
\end{table}

\begin{figure}[tb]
    \centering
    \includegraphics[width=1.0\linewidth]{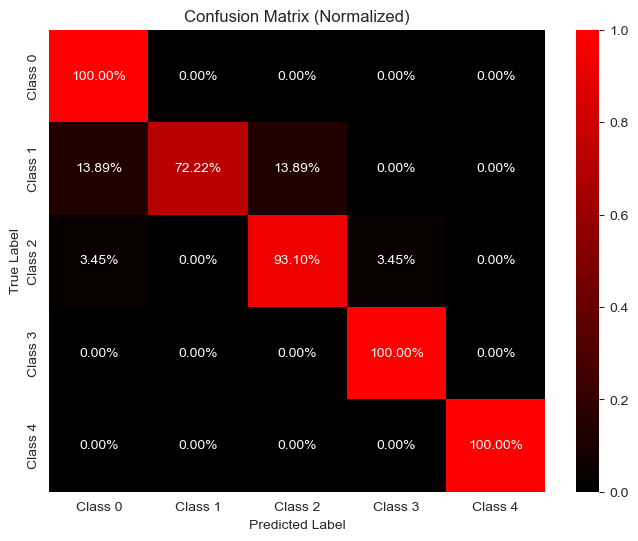}
    \caption{Confusion matrix}
    \label{fig:conf_mat}
\end{figure}
\begin{figure}[tb]
    \centering
    \includegraphics[width=1.0\linewidth]{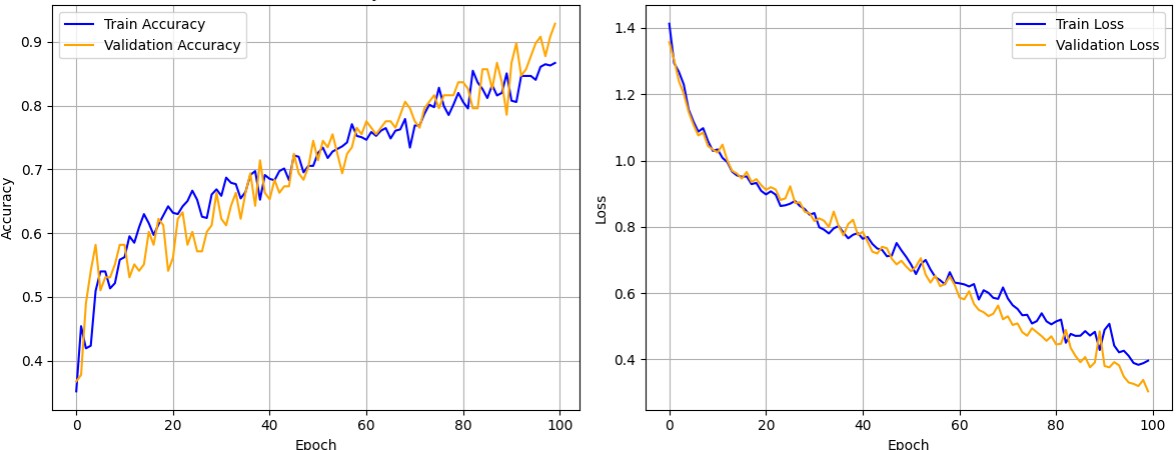}
    \caption{Accuracy and Loss curve}
    \label{fig:conf_mat}
\end{figure}

\subsection{Discussion}
The table compares various regressor models for performance evaluation. Existing works \cite{islam2023using} predominantly use subject-independent approaches, which often limit accuracy due to inter-subject variability. In contrast, the proposed Attention-enhanced CNN-BiLSTM model demonstrates substantially higher performance, highlighting the benefit of integrating attention mechanisms and temporal feature learning for improved prediction.
\begin{table}[htbp]
\centering
\caption{Performance of different regressor models we tested.}
\begin{tabular}{p{3.5cm}|p{1cm}|p{1cm}|p{1cm}|p{1cm}}
\hline
Model & Precision & Recall & F1-Score & Accuracy (\%) \\
\hline
SVR \cite{islam2023using} & N/A & N/A & N/A & 51.94 \\
Random forest regressor \cite{islam2023using} & N/A & N/A & N/A & 49.28 \\
AdaBoost regressor \cite{islam2023using} & N/A & N/A & N/A & 45.60 \\
XGBoost regressor \cite{islam2023using} & N/A & N/A & N/A & 51.33 \\
LightGBM regressor \cite{islam2023using} & N/A & N/A & N/A & 50.92 \\
Shallow neural network-I (one trainable layer) \cite{islam2023using} & N/A & N/A & N/A & 46.83 \\
Shallow neural network-II (two trainable layers) \cite{islam2023using} & N/A & N/A & N/A & 51.33 \\
Proposed Attention-enhanced CNN-BiLSTM & 95.40 & 93.40 & 94.20 & 93.00 \\
\hline
\end{tabular}
\label{tab:regressor_performance}
\end{table}

\section{Conclusion} \label{sec5}.
The proposed hybrid deep learning methodology integrates CNN, BiLSTM, and an attention mechanism to effectively detect multi-class Parkinson’s Disease (PD) based on finger-tapping motion features. By capturing both spatial and temporal patterns, the CNN–BiLSTM framework models intricate hand and wrist movements that are indicative of PD severity. The attention mechanism further enhances the model by focusing on the most informative features within the input sequences, enabling accurate classification across five severity classes. Unlike traditional approaches, this method reduces the reliance on manual interventions and subjective clinical evaluations, offering a more objective and reproducible assessment of disease progression. Using pre-extracted temporal, frequency, and amplitude-based features from finger-tapping videos, the model demonstrates strong performance in distinguishing between different severity levels. This performance highlights the benefit of combining spatial feature extraction, temporal modeling, and attention-driven feature selection within a unified framework. The proposed system provides a non-invasive, efficient, and reliable tool for clinicians to monitor PD progression, supporting timely interventions and personalized patient care. For future work, the model can be extended to incorporate multi-modal data such as gait, speech, and facial expressions, which may further improve classification accuracy and provide a more comprehensive assessment of Parkinson’s Disease severity.

\bibliographystyle{unsrt}
\bibliography{bib-authorship}
\end{document}